\documentclass[10pt,twocolumn,letterpaper]{article}

\usepackage[pagenumbers]{cvpr} 

\usepackage[accsupp]{axessibility}
\usepackage{graphicx}
\usepackage{amsmath}
\usepackage{amssymb}
\usepackage{booktabs}
\usepackage{algorithm}
\usepackage{algorithmic}
\usepackage{array}
\usepackage{xcolor}
\usepackage{url}
\usepackage[misc]{ifsym}
\usepackage{multirow} 
\usepackage{cvpr}
\usepackage{times}
\usepackage{epsfig}
\usepackage{graphicx}
\usepackage{amsmath}
\usepackage{amssymb}
\usepackage{subcaption}
\usepackage{cuted}
\usepackage{capt-of}

\usepackage[pagebackref,breaklinks,colorlinks]{hyperref}

\usepackage[capitalize]{cleveref}
\crefname{section}{Sec.}{Secs.}
\Crefname{section}{Section}{Sections}
\Crefname{table}{Table}{Tables}
\crefname{table}{Tab.}{Tabs.}

\def\cvprPaperID{8964} 
\def\confName{CVPR}
\def\confYear{2023}

\begin{document}

\title{Modeling Entities as Semantic Points for Visual Information Extraction 
\\in the Wild }


\author{Zhibo Yang\textsuperscript{1,2}\thanks{Equal Contribution.} \quad Rujiao Long\textsuperscript{2}\footnotemark[1] \thanks{Correspondence Author.} \quad Pengfei Wang\textsuperscript{2}\footnotemark[1] \quad Sibo Song\textsuperscript{2} \quad Humen Zhong\textsuperscript{2} \quad \\
Wenqing Cheng\textsuperscript{1} \quad Xiang Bai\textsuperscript{1} \quad Cong Yao\textsuperscript{2} \\ 
\textsuperscript{1}Huazhong University  of Science and Technology \qquad
\textsuperscript{2}Alibaba Group 
}
\maketitle

\begin{strip}\centering
\vspace{-16mm}
\includegraphics[width=1.0\textwidth]{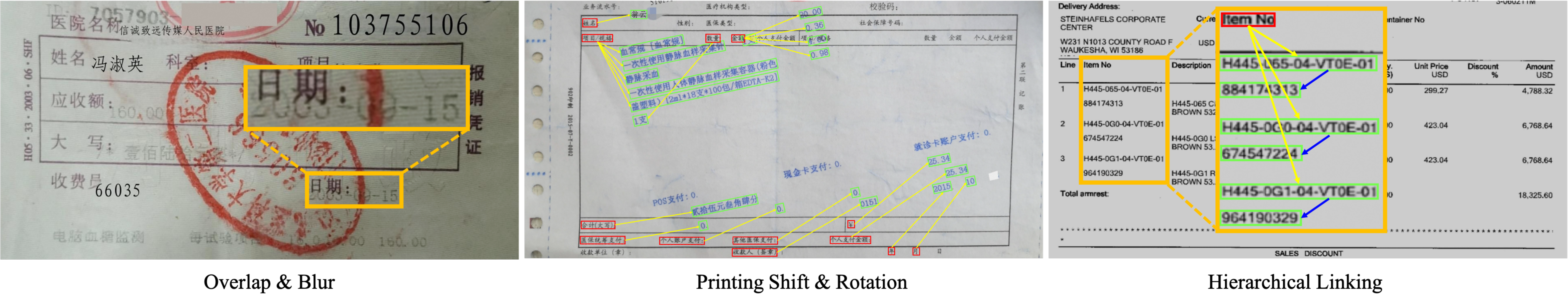}
\vspace{-9mm}
\captionof{figure}{Demonstration of the proposed SIBR dataset, where real-world complexity and challenges (e.g., overlap, blur, printing shift, rotation, and complex linking) are well represented. The rectangles in red and green stand for keys and values, while the yellow and blue arrows indicate inter-entity and intra-entity linkings, respectively. (Best viewed in color.)
\label{fig:demo}}
\vspace{-2mm}
\end{strip}

\newcommand{\zhm}[1]{\textcolor{blue}{[zhm: #1]}}

\begin{abstract}
\vspace{-3mm}
Recently, Visual Information Extraction (VIE) has been becoming increasingly important in both the academia and industry, due to the wide range of real-world applications. Previously, numerous works have been proposed to tackle this problem. However, the benchmarks used to assess these methods are relatively plain, i.e., scenarios with real-world complexity are not fully represented in these benchmarks. As the first contribution of this work, we curate and release a new dataset for VIE, in which the document images are much more challenging in that they are taken from real applications, and difficulties such as blur, partial occlusion, and printing shift are quite common. All these factors may lead to failures in information extraction. Therefore, as the second contribution, we explore an alternative approach to precisely and robustly extract key information from document images under such tough conditions. Specifically, in contrast to previous methods, which usually either incorporate visual information into a multi-modal architecture or train text spotting and information extraction in an end-to-end fashion, we explicitly model entities as semantic points, i.e., center points of entities are enriched with semantic information describing the attributes and relationships of different entities, which could largely benefit entity labeling and linking. Extensive experiments on standard benchmarks in this field as well as the proposed dataset demonstrate that the proposed method can achieve significantly enhanced performance on entity labeling and linking, compared with previous state-of-the-art models. Dataset is available at \url{https://www.modelscope.cn/datasets/damo/SIBR/summary}.
\end{abstract}

\vspace{-4mm}
\section{Introduction} \label{sec:intro}

Visually Rich Documents (VRDs) are ubiquitous in daily, industrial, and commercial activities, such as receipts of shopping, reports of physical examination, product manuals, and bills of entry. Visual Information Extraction (VIE) aims to automatically extract key information from these VRDs, which can significantly facilitate subsequent processing and analysis. Due to its broad applications and grand technical challenges, VIE has recently attracted considerable attention from both the Computer Vision community~\cite{zhang2020trie,wang2022lilt,wang2021towards} and the Natural Language Processing community~\cite{xu2020layoutlm,xu2020layoutlmv2,li2021structurallm}. Typical techniques for tackling this challenging problem include essential electronic conversion of image  (OCR) ~\cite{smith2007overview, Shi2015AnET, Zhou2017EASTAE}, intermediate procedure of structure analysis~\cite{o1993document} and high-level understanding of contents~\cite{xu2020layoutlm}, among which entities play an important role as an aggregation of vision, structure, and language. 

Though substantial progresses~\cite{xu2020layoutlm, li2021structext, huang2022layoutlmv3} have been made, it is still challenging to precisely and reliably extract key information from document images in unconstrained conditions. As shown in Fig.~\ref{fig:demo}, in real-world scenarios documents may have various formats, be captured casually with a mobile phone, or exist occlusion or shift in printing, all of which would pose difficulties for VIE algorithms.

To highlight the challenges in real applications and promote the development of research in VIE, we establish a new dataset called \textbf{S}tructurally-rich \textbf{I}nvoices, \textbf{B}ills and \textbf{R}eceipts in the Wild (\textbf{SIBR} for short), which contains 1,000 images with 71,227 annotated entity instances and 39,004 entity links. The challenges of SIBR lie in: (1) The documents are from different real-world scenarios, so their formats and structures might be complicated and varying; (2) The image quality may be very poor, i.e., blur, noise, and uneven illumination are frequently seen; (3) The printing process is imperfect that shift and rotation might happen.

To deal with these difficulties, we explore an novel approach for information extraction from VRDs. Different from previous methods, which usually employ a sequential pipeline that first uses an off-the-shelf OCR engine to detect and read textual information (location and content) and then fuses such information with visual cues for follow-up entity labeling (\textit{a.k.a.} entity extraction) and linking in a multi-modal architecture (mostly a Transformer)~\cite{xu2020layoutlmv2, li2021structext}, the proposed method adopts a unified framework that all components, including text detection, text recognition, entity extraction and linking, are jointly modeled and trained in an integrated way. This means that in our method a separate OCR engine is no longer necessary. The benefits are two-fold: (1) The accuracy of entity labeling and linking will not be limited by the capacity of the OCR engine; (2) The running speed of the whole pipeline could be boosted.

\begin{figure*}[!t]
  \centering
  \vspace{-2mm}
   \includegraphics[width=0.9\linewidth]{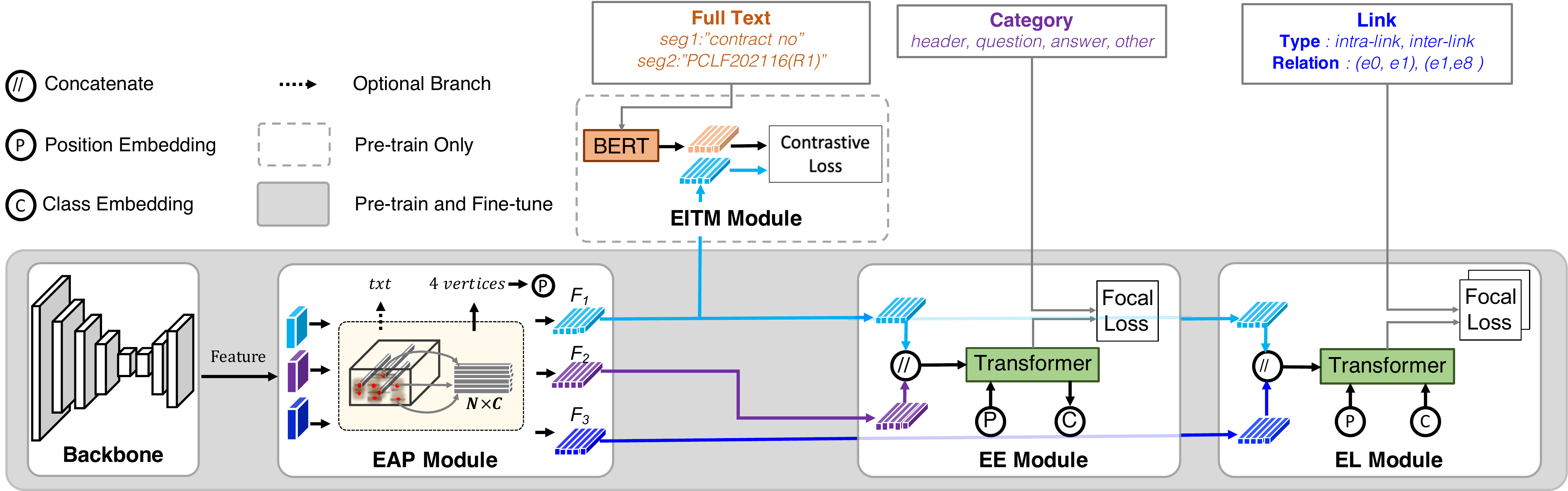}
   \vspace{-3mm}
   \caption{Schematic illustration of the proposed ESP. A ConvNeXt-FPN backbone is used to extract image features. Fed with three convolutional layers appended to the ConvNeXt, EAP module generates three kinds of outputs: (a) Detections, (b) Text contents, (c) Center-based Features: vision-language embedding $F_1$, entity type embedding $F_2$, and entity relational embedding $F_3$. Then, EITM aligns $F_1$ with text embeddings given by BERT by a contrastive learning strategy. In the final decoding stage, EE module takes concatenated features of $F_1$ and $F_2$ as input and predicts categories of the corresponding entities. The process of entity linking (EL) is the same as EE. Note that the recognition branch is optional, as it has little influence on entity extraction, which will be discussed in ablation studies.}
  \label{fig:pipeline}
  \vspace{-5mm}
\end{figure*}

Drawing inspirations from general object detection~\cite{law2018cornernet,zhou2019objects,duan2019centernet,zhou2019bottom} and vision-language joint learning~\cite{pmlr-v139-kim21k, huang2022layoutlmv3, song2022vision}, we put forward to model entities as semantic points (\textbf{ESP} for short). Specifically, as shown in Fig.~\ref{fig:pipeline}, entities are represented using their center points, which are enriched with semantics, such as geometric and linguistic information, to perform entity labeling and linking. To better learn a joint vision-language representation, we also devise three training tasks that are well integrated into the paradigm. The entity-image text matching (\textbf{EITM}) task, which is only used in the pre-training stage, learns to align entity-level vision vectors and language vectors (encoded with off-the-shell BERT) with a contrastive learning paradigm. Entity extraction (\textbf{EE}) and Entity linking (\textbf{EL}), the main tasks for VIE, are used in the pre-training, fine-tuning, and inference stages. In these two modules, region features and position embedding (from ground truth or detection branch) are encoded with transformer layers and then decoded to entity classes and relations. Owing to the joint vision-language representation, text recognition is no longer a necessary module in our framework, and we will discuss the impact of the text recognition branch in Sec.~\ref{sec:Ablation}. 

Extensive experiments have been conducted on standard benchmarks for VIE (such as FUNSD, XFUND, and CORD) as well as the proposed SIBR dataset. We found that compared with previous state-of-the-art methods, the proposed ESP algorithm can achieve highly competitive performance. Especially, it shows an advantage in the task of entity linking. Our main contributions can be summarized as follows: (1) We curate and release a new dataset for VIE, in which the document images are with real-world complexity and difficulties. (2) We devise a unified framework for spotting, labeling and linking entities, where a separate OCR engine is unnecessary. (3) We adopt three vision-language joint modeling tasks for learning informative representation for VIE. (4) Extensive experiments demonstrate the effectiveness and advantage of our approach.

\section{Related Work} \label{sec:relat}

The proposed approach has been inspired by ideas from object detection and document understanding. Exemplar approaches include detection with grouping~\cite{long2021parsing,newell2017associative,xing2023lore}, text spotting~\cite{Zhu2016SceneTD, Long2018SceneTD} and information extraction joint training~\cite{wang2021towards}, and vision-language pre-training for VIE~\cite{xu2020layoutlm}.

\noindent\textbf{Detection with Grouping Methods} consider an object as a collection of parts arranged in specific relations. Object as points methods~\cite{law2018cornernet,zhou2019objects,duan2019centernet} firstly detected four corner points and one center point of an object and then grouped them with the geometric-based algorithm or associative embedding features. SegLink~\cite{shi2017detecting} and its variant SegLink++~\cite{tang2019seglink++} took a small segment of a text instance as the fundamental element and linked them together to form the bounding boxes. In~\cite{hinton1990mapping,hinton2021represent}, hierarchical part-whole ideas were approached at the model level and feature space. Visual entity labeling and linking can also be learned in this manner to some extent. We model visual entities as points and group them with their semantic relations.

\noindent\textbf{Vision-Language Pre-training Methods} usually extract visual document information by using a 2D positional encoding of BERT-like architecture. The innovation of these works lies in novel pre-training tasks or new multi-modal transformer layers. LayoutLM~\cite{xu2020layoutlm} and its variant LayoutLM-v2~\cite{xu2020layoutlmv2} and layoutLM-v3 proposed several novel pretraining tasks, including Masked Visual-Language Model, Text-Image Alignment, Text-Image Matching, and Word-Patch Alignment. StrucText~\cite{li2021structext} introduced a multi-granularity image and text alignment for entity labeling. BROS~\cite{hong2022bros} proposed an area-masking self-supervision strategy that reflected the 2D nature of text blocks. In addition to pre-training tasks, DocFormer~\cite{appalaraju2021docformer} and SelfDoc~\cite{li2021selfdoc} also introduced cross-attention layers for better fusing of visual and language features. LiLT~\cite{wang2022lilt} extended document understanding to multi-languages by designing language-independent attention layers. The above methods usually depended on offline OCR engines, which may lead to error amplification and redundant computation. 

\noindent\textbf{Jointly Training Methods} aim at improving information extraction by a unified end-to-end trainable framework. TRIE~\cite{zhang2020trie} and VIES~\cite{wang2021towards} both proposed networks for simultaneous text detection, text recognition, and information extraction. VIES introduced a vision coordination mechanism and semantics coordination mechanism to gather rich visual and semantic features from text detection and recognition respectively. TRIE designed a multi-modal context block to bridge the OCR and IE modules. Indeed, TRIE and VIES both explicitly classified multi-modal information to get final entity labels without any general visual-language pre-training, but their drawback lies in the poor generalization ability to different downstream tasks. 

\begin{figure}
  \centering
   \includegraphics[width=.85\linewidth]{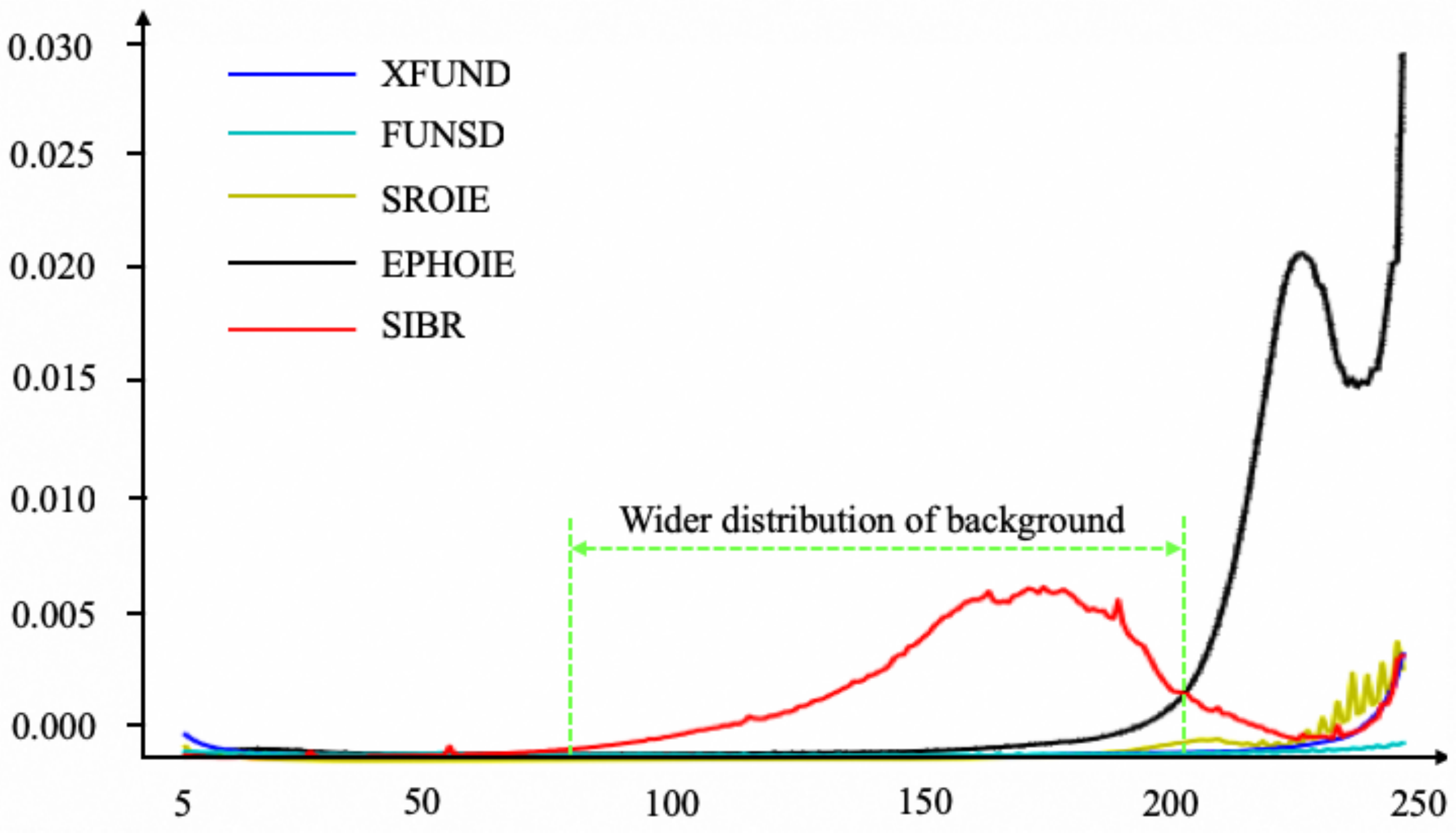}
   \vspace{-0.5\baselineskip}
  \caption{Comparisons of color distribution across different datasets. A distribution of background from 5 to 250 is for better readability. (Best viewed in color.)}
  \label{fig:dataset_rgb_compare}
  \vspace{-2mm}
\end{figure}

\section{The SIBR Dataset}

\begin{table}
\small
\begin{center}
\resizebox{1\linewidth}{!}{
\begin{tabular}{|c|c|c|c|c|c|c|}
\hline
\textbf{Element} & \textbf{SROIE} & \textbf{FUNSD} & \textbf{XFUND} & \textbf{CORD} & \textbf{EPHOIE} & \textbf{SIBR} \\
\hline
Box & 45279 & 9529 & 97012 & 19887 & 15754 & 71227 \\
Link & - & 10420 & 102291 & 7470 & - & 39004 \\
Image & 1139 & 199 & 1393 & 1000 & 1494 & 1000 \\
Scenario & Scan & Scan & Scan & CamC & Scan & CamC \\
Link Type & - & 1 & 1 & 1 & - & 2 \\
Language & 1 & 1 & 7 & 1 & 2 & 2 \\
Overlap & 7.4\% & 9.5\% & 5.1\% & 10.1\% & 1.6\% & 42.7\% \\
\hline
\end{tabular}
}
\end{center}
\vspace{-4mm}
\caption{Comparison between SIBR and other datasets. ``Scan'' is short for scanned receipts or documents, ``CamC'' is short for Camera-captured images. ``Overlap'' showcases the proportion of images with overlapping entity boxes.}
\vspace{-4mm}
\label{table:Dataset_comparsion}
\end{table}

In this section, we introduce \textbf{S}tructurally-rich \textbf{I}nvoices, \textbf{B}ills and \textbf{R}eceipts in the Wild (SIBR) dataset.

\noindent\textbf{Dataset Description} With the wide spread of mobile phones, photoed document images are ubiquitous in industrial and commercial activities, among which photoed receipts are of great value for their probative force to business transactions. There are 1000 images in the SIBR, including 600 Chinese invoices, 300 English bills of entry, and 100 bilingual receipts. SIBR is well annotated with 71227 entity-level boxes and 39004 links. The dataset is divided into a training set of 600 images and a test set of 400 images. Details are shown in Table.~\ref{table:Dataset_comparsion}.

\noindent\textbf{Dataset Challenge} Previously, several datasets~\cite{jaume2019funsd, park2019cord} have been constructed for VIE. However, they are relatively plain, i.e., scenarios with real-world complexity are not well represented in these benchmarks. Compared with real scene datasets such as (SROIE~\cite{huang2019icdar2019} and EPHOIE~\cite{wang2021towards}), SIBR has more diverse appearances and richer structures. The document images in SIBR are much more challenging in that they are taken from real applications and difficulties (Fig.~\ref{fig:demo}) such as severe noise, uneven illumination, image deformation, printing shift and complicated links. As shown in Fig.~\ref{fig:dataset_rgb_compare}, SIBR has a wider distribution of RGB value. Due to image deformation or printing shift, SIBR has 42.7\% of images with overlapping boxes, as shown in the bottom line of Table.~\ref{table:Dataset_comparsion}, which brings great challenges to text spotting and information extraction.

\noindent\textbf{Annotation Details} Following ~\cite{wang2021towards}, four vertices are used to represent the bounding box of the entity. The label of entity is categorized as ``question'', ``answer'', ``header'', and ``other''. The link between two entities is categorized as either an intra-link or an inter-link, which represents links in an entity or between two different entities. In addition, text content is also annotated for end-to-end information extraction. It is worth noting that the granularity of text boxes is different in SIBR and FUNSD~\cite{jaume2019funsd}. FUNSD is annotated at the semantic block level which may contain multiple lines, while the entity with multiple lines in SIBR is represented by text segments and intra-links between segments. As shown in  Fig.~\ref{fig:demo}, the blue lines represent intra-entity links, while the yellow lines represent inter-entity links. With the help of ~\cite{yu2019free}, the sensitive information (names, telephone numbers, etc.) is erased and re-edited with new text.

\section{Methodology}

In this section, we introduce our ESP framework for visual information extraction tasks, including entity extraction (EE) and entity linking (EL). 
The proposed ESP is first pre-trained with pseudo-labels and objectives which are carefully designed to learn entity semantics as well as the entity relations. Then we fine-tune our model for downstream tasks using task-specific annotations.

\subsection{Overall Architecture}
The goal of a VIE algorithm is to learn to locate key text contents and extract entity labels and linking information from visually-rich documents. 
We approach VIE from a point-based perspective, i.e., we learn to represent each entity, its label and relations with other entities by a single point. Specifically, we adopt the CenterNet framework~\cite{zhou2019objects}, owing to its simple and concise pipeline. By utilizing the unified paradigm for feature learning, the model can benefit from joint learning across different modalities and tasks.

\noindent\textbf{Visual Backbone}
We choose ConvNeXt-tiny-FPN as the visual backbone for its excellent trade-off between performance and efficiency. ConvNeXt~\cite{liu2022convnet} is a pure CNN architecture, which compares favorably with state-of-the-art hierarchical vision Transformers across multiple vision benchmarks while retaining the simplicity and efficiency of ConvNets. Moreover, we adopt feature pyramid network (FPN)\cite{lin2017feature} to fully exploit multi-scale visual features. Let $I \in \mathbb{R}^{w \times h \times 3}$ be the input image of width $w$ and height $h$. 
The visual backbone then produces a feature map $P_2 \in \mathbb{R}^{\frac{w}{4} \times \frac{h}{4} \times 256}$ as the input to the EAP module.

\noindent\textbf{Entities-As-Points (EAP)}
Given an image, CenterNet~\cite{zhou2019objects} extracts a set of bounding boxes for each object class with the help of a heatmap head, a size head and an offset head. We build the EAP module upon the original CenterNet framework with the following architectural modifications. 

First, we devise the size head into a quadrangle head, which predicts the location of four vertices to detect arbitrary-oriented entity text instances. Therefore, the total text detection loss is computed as follows:
\begin{equation}
  \mathcal{L}_{\text{Det}} = \mathcal{L}_{\text{Heatmap}} + \mathcal{L}_{\text{Quad}} + \mathcal{L}_{\text{Offset}}
\end{equation}
Second, we propose three feature transformation branches to encode cross-modal features, entity type features, and entity relational features respectively. Each feature transformation branch consists of a $3\times3$ convolutional layer, a ReLU layer, and a final $1\times1$ convolutional layer. Moreover, the transformed feature maps are extracted based on the predicted points from the text detection branch. 
In this way, cross-modal features and entity-related features are produced in a unified point-based representation. Third, we adopt a simplified version of CRNN~\cite{shi2016end} as the text recognition branch. The predicted entity boxes will be fed into a grid-sample module to sample the text features $F_t \in \mathbb{R}^{N \times C \times S_h \times S_w}$. $S_h$ and $S_w$ denote the sampling size of height and width respectively while N denotes the number of text segments. Our CRNN variant is composed of 6 convolutional layers for feature extraction, 1 bi-directional LSTM layer for sequential modeling, and 1 linear projection layer for label prediction. Connectionist Temporal Classification (CTC) loss~\cite{graves2006connectionist} is adopted as recognition loss $\mathcal{L}_{\text{Rec}}$ to align sequence predictions with the labels. In fact, our model does not depend on the recognition module, which is only used to output the end-to-end results, so we did not use the recognition results on FUNSD, XFUND, and CORD.

\vspace{-4mm}
Overall, the point-based nature of entity representation possesses several advantages. 
First, our proposed method learns visual, textual, and layout representations simultaneously on a grid-based feature map, avoiding sophisticated serialization steps as for sequence-based methods~\cite{xu2020layoutlm,xu2021layoutxlm,huang2022layoutlmv3}, especially in real-world scenarios (e.g., misaligned second-time printing, receipts with complex layout) where such linearization step is nontrivial. 
Second, the text is usually truncated with a preset maximum number of tokens for sequence-based approaches. Nevertheless, our proposed model is not limited by a token number and can learn more holistic representations to mine inter-text relationships without any text truncation.

\noindent\textbf{Entity-Image Text Matching (EITM)}
Recent progress in VIE has been mainly powered by multi-modal pre-training. Therefore, we harness an entity-image text matching module during the pre-training stage to enhance cross-modal feature learning. An image-text matching task is adopted to encourage cross-modal interactions and mutual alignment between visual and textual representations. EITM is only applied to the pre-training stage, thus enabling recognition-free prediction for entity labeling and linking. We employ the idea of contrastive learning to mutually align entity image and entity text representations into semantic space. It aims to find the closest entity image embedding from a batch given an entity text embedding. Similarly, for a given entity image embedding, the objective is to find the closest entity text embedding. This results in an InfoNCE loss over each batch of entity image-text pairs for pre-training our model,

\begin{equation}
  L_{v2t} = - \sum_{j=1}^{N} \log \frac{\exp(v_i \cdot t_j / \tau)}{\sum_{k=1}^{N} \exp(v_i \cdot t_k / \tau)}
  \label{eq:v2t}
\end{equation}
where $N$ denotes the number of image-text pairs which equals the number of text segments, $v_i$ and $t_i$ represent $i$-th entity image embedding and entity text embedding respectively, and $\tau$ denotes the hyperparameter of temperature. Similarly, for each $t_i$, the loss is formulated as,

\vspace{-2mm}
\begin{equation}
  L_{t2v} = - \sum_{j=1}^{N} \log \frac{\exp(t_i \cdot v_j / \tau)}{\sum_{k=1}^{N} \exp(t_i \cdot v_k / \tau)}
  \label{eq:t2v}
\end{equation}

Finally, the total loss for EITM is defined as,
\vspace{-2mm}
\begin{equation}
  L_{\text{EITM}} = L_{v2t} + L_{t2v}
  \label{eq:citm}
\end{equation}

\noindent\textbf{Entity Extraction (EE)}
To strengthen the entity type feature learning, we first concatenate vision-language features $F_1$ and entity type features $F_2$ to leverage both visual and semantic representations. 

We then employ two self-attention layers to aggregate features from all entities. Last, the aggregated features are fed into the entity classifier, which is a ReLU-fc-ReLU-fc-Sigmoid block. We set the hidden size, number of self-attention heads, feed-forward/filter size, and number of transformer layers to 832, 8, 64, and 2 respectively. The dimension of classifier output is $N \times N_c$, where $N$ denotes the number of text segments and $N_c$ denotes the number of entity classes.  

To address the extreme imbalance of entity types, we adopt Focal loss~\cite{lin2017focal} as our loss function,

\vspace{-2mm}
\begin{equation}
  L_{\text{EE}} = \left\{
  \begin{aligned}
  - (1-p_{t}) ^\gamma \log (p_{t}) , \quad p_{t}^* = 1 \\
  -p_{t}^\gamma \log (1-p_{t}) , \quad   otherwise
  \end{aligned}
  \right.
  \label{eq:ctl}
\end{equation}
where $p_{t}$ is the predicted probability vector and $p_{t}^*$ is the ground truth. We set $\gamma$ to 4.

\noindent\textbf{Entity Linking (EL)}
In many VIE applications, an entity often contains multiple text segments or text boxes. Therefore, it is important to model intra-entity links in addition to inter-entity links. Inter-entity links rely more on the spatial correlations of text segments, however, intra-entity links rely heavily on text semantics. 
Thus, we concatenate the cross-modal feature $F_1$ with entity relational feature $F_3$ to handle the lack of semantics and apply two layers' self-attention modules to capture inter-text dependencies.

The structure of the module is the same as Entity Extraction Module. 
The only difference is that the Entity Linking Module constructs a relationship matrix between entity text instances. 
We denote the output of the self-attention layers by $E_{l}^{'} \in \mathbb{R}^{N_{l} \times C}$, where $C$ is the dimension of input feature embeddings and $N_{l}$ is the number of key-value pairs.
First, we broadcast $E_{l}^{'}$ to obtain $E_{l1} \in \mathbb{R}^{1 \times N_{l} \times d}$ and $E_{l2} \in \mathbb{R}^{N_{l} \times 1 \times d}$. 
Since entity links are vectors with directions, we formulate the relational matrix via element-wise subtraction as given in Eq.~\eqref{eq:nxn} to encourage learning directed linking representations, 

\begin{equation}
  E_{l}^{*} = E_{l1} - E_{l2}, \quad E_{l}^{*} \in \mathbb{R}^{N_{l} \times N_{l} \times C}
  \label{eq:nxn}
\end{equation}
$E_{l}^{*}$ is input into the classifier to obtain link results. The classifier is a ReLU-fc-ReLU-fc-Sigmoid block. 
Since the entity linking matrices are usually sparse, we use Focal loss as the same as the entity labeling task and denote as $L_{\text{EL}}$.

\subsection{Training Objectives}
\noindent\textbf{Pre-Training} 
The pre-training is conducted on DocBank and RVL-CDIP datasets with all aforementioned modules. Unlike previous methods that focus on designing mask-based or cross-alignment tasks, we generate pseudo-labels to mine semantic and relational information in these pre-training datasets. See \cref{paragraph:dataset_construction} for more details. Thus, the overall pre-training objective is,
\vspace{-2mm}
\begin{equation}
  \mathcal{L}_{\text{PT}} = \mathcal{L}_{\text{Det}} + \mathcal{L}_{\text{EITM}} + \lambda_{EL}\mathcal{L}_{\text{EL}} + \lambda_{EE}\mathcal{L}_{\text{EE}}
\end{equation}

\noindent\textbf{Fine-Tuning} Since EITM task only exists in the pre-training stage and recognition tasks only exist in fine-tuning, the loss term for fine-tuning is,
\vspace{-2mm}
\begin{equation}
  \mathcal{L}_{\text{FT}} = \mathcal{L}_{\text{Det}} + \mathcal{L}_{\text{Rec}}^* + \lambda_{EL}\mathcal{L}_{\text{EL}} + \lambda_{EE}\mathcal{L}_{\text{EE}}
\end{equation}
where $\lambda_{EL}$ is set to 1 for pre-training and 10 for fine-tuning and $\lambda_{EE}$ is set to 10 for pre-training and 200 for fine-tuning. 
It is worth noting that, the recognition branch in our framework is optional, i.e., it can be removed or replaced by any off-the-shelf text recognizer given entity images.

\begin{table*}
\small
\begin{center}
\vspace{-3mm}
\begin{tabular}{|c|c|c|p{1.2cm}<{\centering}|p{1.2cm}<{\centering}|p{1.2cm}<{\centering}|p{1.2cm}<{\centering}|p{1.2cm}<{\centering}|p{1.2cm}<{\centering}|}
\hline
\multirow{2}{*}{\textbf{Model}} & \multirow{2}*{\textbf{Parameters}} & \textbf{Pretrain} &
\multicolumn{2}{c|}{\textbf{FUNSD}} &\multicolumn{2}{c|}{\textbf{XFUND}} &\multicolumn{2}{c|}{\textbf{CORD}}\\
\cline{4-9}
& & \textbf{Data Size} & \textbf{EE} & \textbf{EL} & \textbf{EE} & \textbf{EL} & \textbf{EE} & \textbf{EL} \\
\hline
$\text{BERT}_\text{BASE}$~\cite{xu2020layoutlm} & 110M + $\alpha$ & - & 60.92 & 27.65 & - & - & 89.68 & 92.83 \\
$\text{RoBERTa}_\text{BASE}$~\cite{cui2020revisiting} & 125M + $\alpha$ & - & 66.48 & - & - & 47.69 & 93.54 & - \\
$\text{LayoutLM}_\text{BASE}$~\cite{xu2020layoutlm} & 160M + $\alpha$ & 11M & 88.41 & 45.86 & - & - & 96.07 & 95.21 \\
SelfDoc~\cite{li2021selfdoc} & - & 0.32M & 83.36 & - & - & - & - & - \\
UDoc~\cite{gu2021unidoc} & 272M + $\alpha$ & 11M & 87.93 & - & - & - & - & - \\
StrucText~\cite{li2021structext} & 107M + $\alpha$ & 0.9M & 83.09 & 44.10 & - & - & - & - \\
$\text{LayoutLMv2}_\text{BASE}$~\cite{xu2020layoutlmv2} & 200M + $\alpha$ & 11M & 82.76 & 42.91 & - & - & 94.95 & 95.59 \\
$\text{LiLT}_\text{BASE}$~\cite{wang2022lilt} & - & 11M & 85.74 & 62.76 & 85.85 & 81.25 & - & - \\
$\text{XYLayout}_\text{BASE}$~\cite{gu2022xylayoutlm} & 345M + $\alpha$ & 30M & 83.35 & - & 82.04 & 67.79 & - & - \\
$\text{BROS}_\text{BASE}$~\cite{hong2022bros} & 110M + $\alpha$ & 11M & 83.05 & 71.46 & - & - & 96.50 & 95.73 \\ 
$\text{LayoutLMv3}_\text{BASE}$~\cite{huang2022layoutlmv3} & 133M + $\alpha$ & 11M & 90.29 & - & - & - & 96.56 & \\
\hline
$\text{BERT}_\text{LARGE}$~\cite{xu2020layoutlm} & 340M + $\alpha$ & - & 65.63 & 29.11 & - & - & 90.25 & 94.31 \\
$\text{RoBERTa}_\text{LARGE}$~\cite{cui2020revisiting} & 355M + $\alpha$ & - & 70.72 & - & - & - & 93.80 & - \\
$\text{LayoutLM}_\text{LARGE}$~\cite{xu2020layoutlm} & 343M + $\alpha$ & 11M & 77.89 & 42.83 & - & - & - & 95.41 \\
$\text{LayoutLMv2}_\text{LARGE}$~\cite{xu2020layoutlmv2} & 426M + $\alpha$ & 11M & 84.20 & 70.57 & - & - & 96.01 & 97.29 \\
$\text{BROS}_\text{LARGE}$~\cite{hong2022bros} & 340M + $\alpha$ & 11M & 84.52 & 77.01 & - & - & 97.28 & 97.40 \\
$\text{DocFormer}_\text{LARGE}$~\cite{appalaraju2021docformer} & 536M + $\alpha$ & 5M & 84.55 & - & - & - & 96.99 & - \\
$\text{LayoutLMv3}_\text{LARGE}$~\cite{huang2022layoutlmv3} & 368M + $\alpha$ & 11M & \textbf{92.08} & 80.35 & - & - & \textbf{97.46} & 98.28 \\
\hline
ESP (ours)  & 50M & 0.9M & 91.12 & \textbf{88.88} & \textbf{89.13} & \textbf{92.31} & 95.65 & \textbf{98.80} \\
\hline
\end{tabular}
\end{center}
\vspace{-5mm}
\caption{Comparisons on Entity Extraction and Entity Linking tasks. $\alpha$: parameters for OCR engine should be considered. }
\label{comparing_SOTA}
\vspace{-5mm}
\end{table*}

\section{Experiments}

\begin{figure}[]
  \centering
   \includegraphics[width=0.8\linewidth]{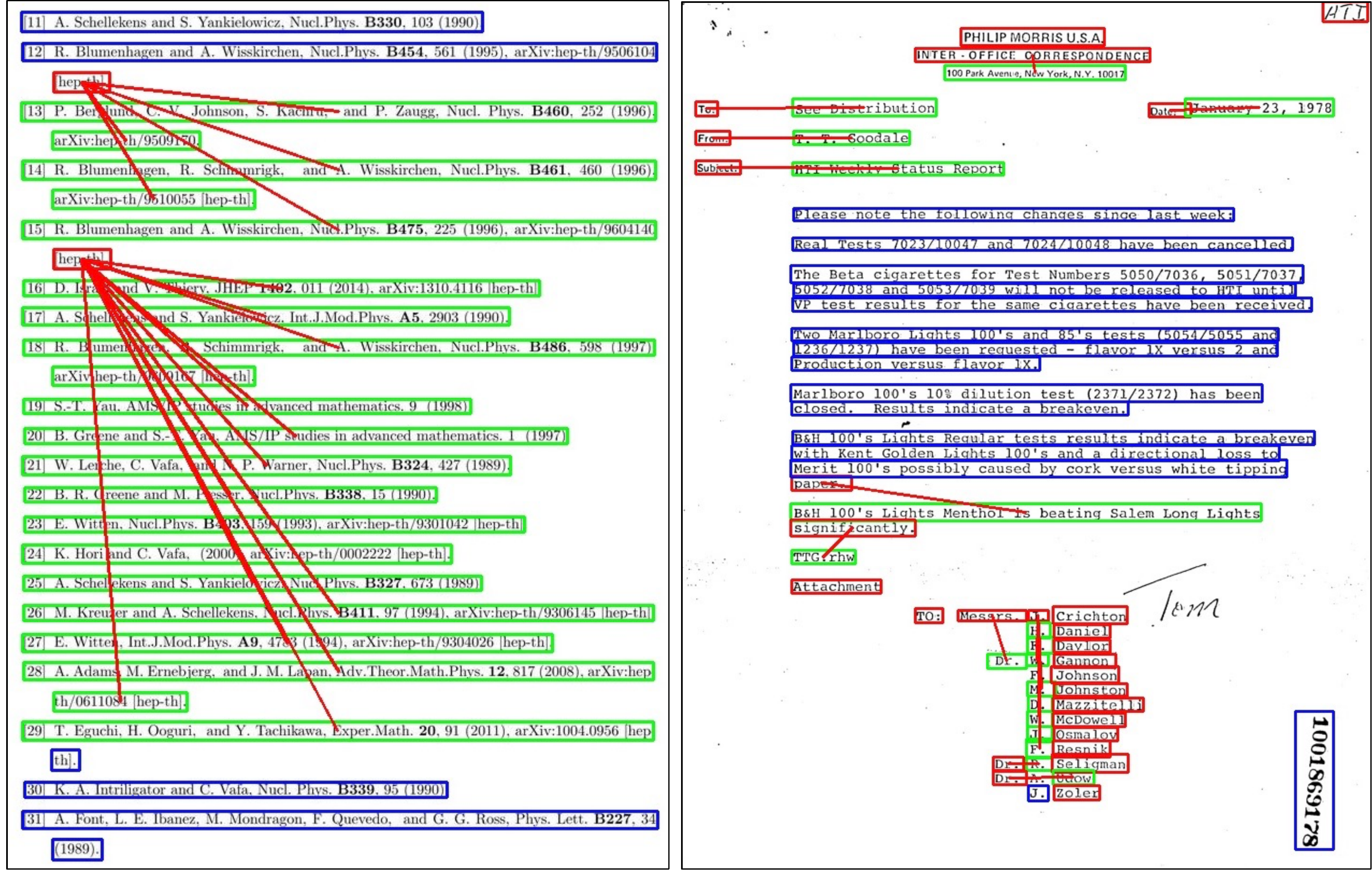}
   \vspace{-3mm}
   \caption{Example of pseudo labels. The boxes in red, green, and blue represent pseudo keys, pseudo values, and unmatched text contents, respectively. Key-value pairs are linked with red lines.}
  \label{fig:Generated pseudo labels}
  \vspace{-5mm}
\end{figure}

\subsection{Datasets}
DocBank and RVL-CDIP are used for pre-training, while the FUNSD~\cite{jaume2019funsd}, XFUND~\cite{wang2021towards} and CORD~\cite{park2019cord} are adopted to compare with SOTA methods to verify the effectiveness of the ESP on EE and EL tasks. 

Following ~\cite{li2021structext}, we use both \textbf{DocBank} and \textbf{RVL-CDIP} for pre-training. DocBank contains 500K images, of which 400K are for training, 50K for validation, and 50K for testing. RVL-CDIP includes 400K grayscale images in 16 categories, with 25K images per category. 320K are for training, 40K for validation, and 40K for testing. 

\textbf{FUNSD} is composed of 149 training samples and 50 testing samples, where the images are scanned English documents. \textbf{XFUND} extends the FUNSD dataset to 7 other languages, with the same number of images in each language as the FUNSD. \textbf{CORD} consists of 1000 fully annotated receipts, including 800 training samples, 100 validation samples, and 100 testing samples. The FUNSD, XFUND, and CORD are annotated with both word and block locations. Combining coordinate information of text words and blocks, the segment-level annotation is generated for fine-tuning the ESP model.

\noindent\textbf{Pre-Training Pseudo Label Generation}
\label{paragraph:dataset_construction}
The existing pre-training methods mainly focus on designing self-supervised tasks to boost performance, such as masked language models or word-patch alignment. We approach this by exploiting knowledge from pre-training datasets. The pipeline to generate pseudo-labels for pre-training EE and EL tasks of key-value pairs is described as follows:

First, the key library is generated by counting the occurrences of text contents. The number of occurrences more than 100 and 15 will be considered as key in DocBank and RVL-CDIP, respectively. Generally, the keys should not be too long or too short, thus the text, with more than 50 characters or less than 3, is skipped. Second, the key-value pairs are generated based on the spatial relationship. If a text appears in the key library, it will be considered as a key, and then we can find its corresponding values based on rules, such as the values will appear below or right of the keys empirically. The category of other text is assigned as ``other". For the key-value pairs, the key/value attributes are actually the pseudo labels of the EE task, and the key-value relationships are the pseudo labels of the EL task.

The visualization of pseudo-label is shown in Fig.~\ref{fig:Generated pseudo labels}. For there are some errors, the EL task is only trained with matched key/value links and the others are masked in the pre-training. 

\subsection{Implementation Details}

\noindent\textbf{Pre-Training} We pre-train the ESP models for 10 epochs with an initial learning rate of 1.25e-4, which is decayed by a factor of 0.1 at the 5th and 8th epoch, respectively. The longer side of images is resized to 1024, with the aspect ratio kept. The OCR information, obtained from PaddleOCR~\cite{Li2022PPOCRv3MA}, is adopted to generate pseudo labels, including key-value labeling and linking information with the algorithm mentioned in \cref{paragraph:dataset_construction}.

\noindent\textbf{Fine-Tuning} Keeping the input size the same as the pre-training stage, the batch size is set to 8 and the initial learning rate is 1.25e-4. Since there are only 149 training samples in FUNSD, we train our model for 300 epochs, and the learning rate drops to 1.25e-5 and 1.25e-6 at the 200th and 250th epochs, respectively. CORD and XFUND are trained for 150 epochs, and the learning rate drops to 1.25e-5 and 1.25e-6 at the 100th and 130th epochs. 

\noindent\textbf{Evaluation} Following~\cite{li2021structext}, we aggregate segment features of the text into one entity if the entity has multiple segments on FUNSD, XFUND, and CORD. But when testing end-to-end on SIBR, model inference does not rely on any annotation information. After decoding, the boxes, categories, and links are directly matched with GT. We use F1-score as our evaluation metrics for both EE and EL tasks.

\begin{table}
\small
\begin{center}
\resizebox{1\linewidth}{!}{
\begin{tabular}{|c|c|c|c|c|c|c|c|c|c|c|}
\hline
\multirow{2}{*}{\textbf{Task}} & \multirow{2}*{\textbf{Model}} & \multirow{2}*{\textbf{FUNSD}} &
\multicolumn{8}{c|}{\textbf{XFUND}} \\
\cline{4-11}
& & & \textbf{ZH} & \textbf{JA} & \textbf{ES} & \textbf{FR} & \textbf{IT} & \textbf{DE} & \textbf{PT} & \textbf{Avg.} \\
\hline
\multirow{4}{*}{EE} & $\text{RoBERTa}$~\cite{cui2020revisiting} & 66.7 & 87.7 & 77.6 & 61.1 & 67.4 & 66.9 & 68.1 & 68.2 & 71.0 \\
& $\text{LayoutXLM}$~\cite{xu2021layoutxlm} & 79.4 & 89.2 & 79.2 & 75.5 & 70.2 & 80.8 & 82.2 & 79.0 & 79.5 \\
& $\text{LiLT}$~\cite{wang2022lilt} & 84.2 & 89.4 & 79.6 & 79.1 & 79.5 & 83.7 & 82.3 & 82.2 & 82.3 \\
& ESP & \textbf{91.1} & \textbf{90.3} & \textbf{81.1} & \textbf{85.4} & \textbf{90.5} & \textbf{88.9} & \textbf{87.2} & \textbf{87.5} & \textbf{87.3} \\
\hline
\multirow{4}{*}{EL} & $\text{RoBERTa}$~\cite{cui2020revisiting} & 26.6 & 51.1 & 58.0 & 52.9 & 49.7 & 53.1 & 50.4 & 39.8 & 50.7 \\
& $\text{LayoutXLM}$~\cite{xu2021layoutxlm} & 54.8 & 70.7 & 69.6 & 68.9 & 63.5 & 64.2 & 65.5 & 57.2 & 65.6 \\
& $\text{LiLT}$~\cite{wang2022lilt} & 62.76 & 72.9 & 70.4 & 71.9 & 69.7 & 70.4 & 65.6 & 58.7 & 68.5 \\
& ESP & \textbf{88.6} & \textbf{90.8} & \textbf{88.3} & \textbf{85.2} & \textbf{90.9} & \textbf{90.0} & \textbf{85.2} & \textbf{86.2} & \textbf{88.1} \\
\hline
\end{tabular}
}
\end{center}
\vspace{-4mm}
\caption{Evaluation on FUNSD and XFUND (fine-tuning on X, testing on X). Note that, LayoutXLM adopts language-specific fine-tuning and LiLT applies a plain text model (InfoXLM). However, ESP does not involve any multilingual language models as EITM is frozen during the fine-tuning stage. }
\vspace{-5mm}
\label{multi-language}
\end{table}

\begin{figure*}[h]
\centering
\vspace{-3mm}
\includegraphics[width=0.79\textwidth]{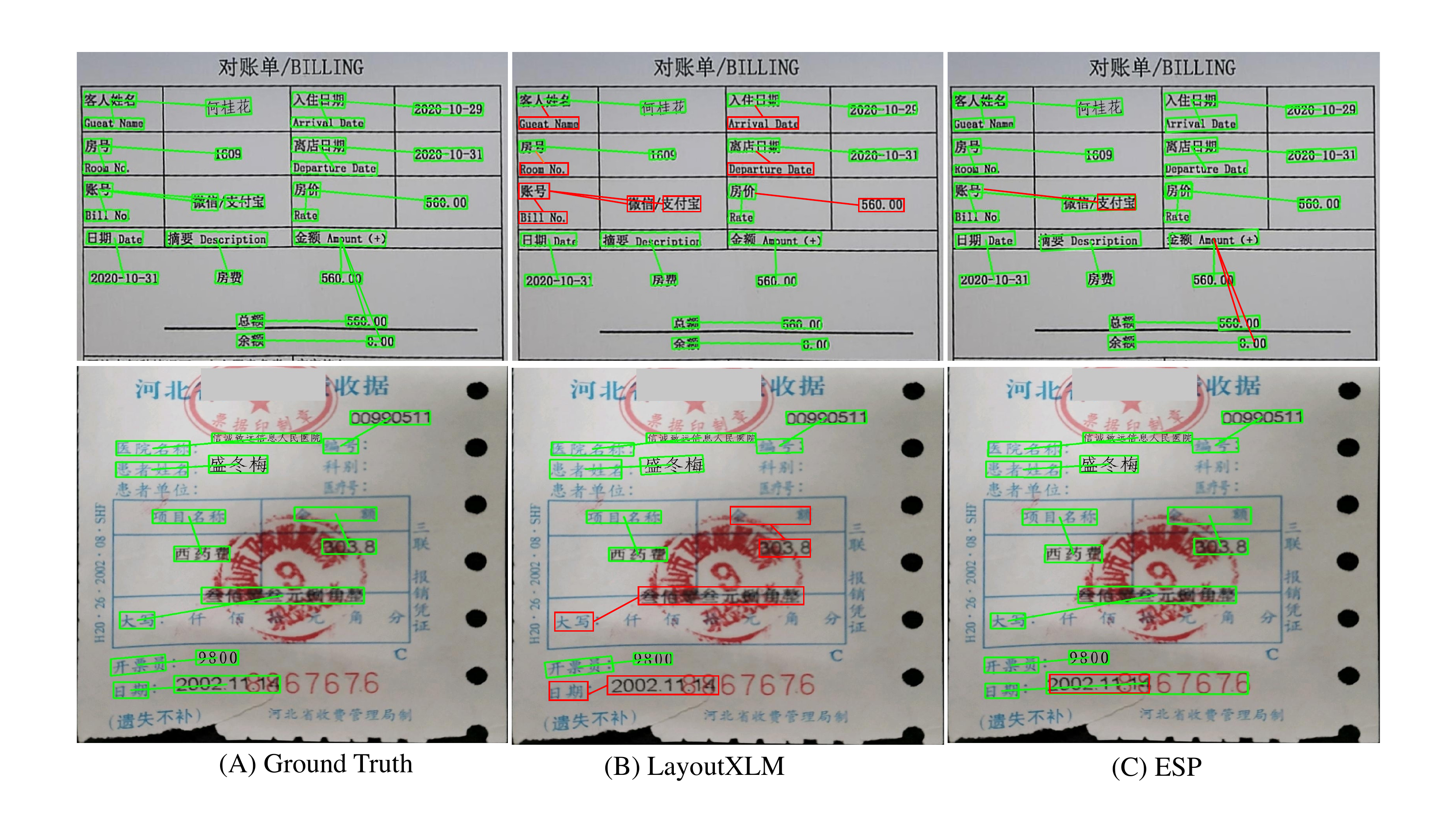}
\vspace{-2mm}
\caption{Qualitative results of ESP on the proposed dataset. The columns from left to right represent the visualization of ground truth, LayoutXLM, and ESP. The rectangles in green and red stand for the results of correct and incorrect information extraction, respectively. The LayoutXLM is more error-prone, due to the lower quality of OCR results. Only key-value pairs are shown for better visualization.}
\label{fig:final_vis}
\vspace{-5mm}
\end{figure*}

\subsection{Comparison with the State-of-the-Arts}

In this section, we compare ESP with LayoutLM-based models~\cite{xu2020layoutlm,xu2020layoutlmv2, xu2021layoutxlm, huang2022layoutlmv3}, BERT~\cite{devlin2018bert}, RoBERTa~\cite{cui2020revisiting}, XYLayout~\cite{gu2022xylayoutlm}, StrucText~\cite{li2021structext}, and BROS~\cite{hong2022bros} on three public benchmarks, and the results are summarized in ~\cref{comparing_SOTA}.

We observe that ESP outperforms all existing methods on the EL task, and it boosts the upper bound of FUNSD and XFNSD to new record (88.88\% and 92.31\%). This proves that visual and geometric features can largely benefit entity linking prediction. On the EE task, ESP also achieves competitive performance, but its performance is slightly lower than that of LayoutLMv3 on FUNSD and CORD. Note that~\textit{during evaluation LayoutLMv3 additionally used the ground truth of text contents, while ESP did not, so the comparison is not totally fair}. Moreover, the model size (50M parameters) and the image set for pre-training (0.9M samples) of ESP are much smaller, compared with LayoutLMv3 (over 368M parameters and 11M samples).

We also conduct experiments on XFUND and FUNSD to verify the importance of visual features, and the results are shown in Tab.~\ref{multi-language}. Previous LayoutLM and its variants extract image feature by additional un-trainable visual encoders, leading to inevitably losses of OCR related details. We argue that fine-grained visual cues, such as fonts, colors, positions and typographic information, are crucial for VIE. We compare ESP with three multi-language approaches, LayoutXLM, RoBERTa, and LiLT. We pre-train ESP on English datasets of DocBank and RVL-CDIP, but fine-tune and test it on language-specifically FUNSD (English) and XFUND (7 other languages). Note that the EITM task is not trained in the fine-tuning stage, meaning that the model does not perceive the other 7 kinds of languages. We find that ESP maintains at a plateau of 87.3\% on XUFNSD, much better than language-independent LiLT, which demonstrates the effectiveness of visual information extractors.

\begin{table}
\small
\begin{center}
\begin{tabular}{|c|c|c|c|c|c|c|c|}
\hline
\multirow{2}{*}{\textbf{Tag}} & 
\multicolumn{4}{c|}{\textbf{Task}} &\multicolumn{3}{c|}{\textbf{FUNSD}} \\
\cline{2-8}
& \textbf{EAP} & \textbf{EITM} & \textbf{EE} & \textbf{EL} & \textbf{EE} & \textbf{EL} & \textbf{Box}  \\
\hline
\hypertarget{T4a}{(a)} & \checkmark & \checkmark & \checkmark & \checkmark & \textbf{91.12} & \textbf{88.89} & \textbf{78.00} \\
\hypertarget{T4a}{(b)} &  & \checkmark & \checkmark & \checkmark & 89.86 & 88.49 & 70.07 \\
\hypertarget{T4a}{(c)} & \checkmark & \checkmark &  & \checkmark & 88.60 & 87.97 & 77.85 \\
\hypertarget{T4a}{(d)} & \checkmark &  & \checkmark & \checkmark & 88.67 & 88.44 & 76.87 \\
\hypertarget{T4a}{(e)} & \checkmark & \checkmark & \checkmark &  & 89.13 & 61.20 & 77.05 \\
\hypertarget{T4a}{(f)} & \checkmark &  &  &  & 85.10 & - & 73.65 \\
\hypertarget{T4a}{(g)} &  &  &  &  & 85.04 & 60.45 & 66.09 \\
\hline
\end{tabular}
\end{center}
\vspace{-4mm}
\caption{Ablation study for pre-training tasks. The segment-level boxes are generated according to the locations of the words. However, the results are not high due to inaccurate word locations.} 
\label{ablation}
\vspace{-4mm}
\end{table}

\begin{table}
\small
\begin{center}
\resizebox{1\linewidth}{!}{
\begin{tabular}{|c|c|c|c|c|}
\hline
\textbf{Tag}&\textbf{Task} & \textbf{Model} & \textbf{EE} & \textbf{EL} \\
\hline
\multirow{3}{*}{(a)} & \multirow{3}{*}{EE/EL} & 
TRIE~\cite{zhang2020trie} & 85.62 & - \\ &
& LayoutXLM~\cite{xu2021layoutxlm} & 94.72 & 83.99 \\
& & $\text{ESP}^{\dagger}$ & 94.93 & 85.87 \\
& & ESP & \textbf{95.27} & \textbf{85.96} \\
\hline
\multirow{3}{*}{(b)} & \multirow{3}{*}{Det + EE/EL} & 
TRIE~\cite{zhang2020trie} & 46.40 & - \\ &
& LayoutXLM~\cite{xu2021layoutxlm} & 75.07& 53.63 \\
& & $\text{ESP}^{\dagger}$ & 91.69 & 73.22 \\
& & ESP & \textbf{92.29} & \textbf{74.44} \\
\hline
\multirow{3}{*}{(c)} & \multirow{3}{*}{Det + Rec + EE/EL} & 
TRIE~\cite{zhang2020trie} & - & - \\ &
& LayoutXLM~\cite{xu2021layoutxlm} & 68.55 & 46.71 \\
& & $\text{ESP}^{\dagger}$ & 67.76 & 48.17 \\
& & ESP & \textbf{70.45} & \textbf{51.47} \\
\hline
\end{tabular}
}
\end{center}
\vspace{-5mm}
\caption{Evaluations and ablation studies on SIBR. ``Det'' and ``Rec'' denote ``detection'' and ``recognition'', respectively. The recognition of text is considered correct when the normalized edit distance is less than 0.3. The $\dagger$ denotes a co-training process with text recognition branch. }
\label{res_on_SIBR}
\vspace{-5mm}
\end{table}

\subsection{Ablation Study}\label{sec:Ablation}

\noindent\textbf{Effect of Pre-training Tasks} This group of experiments is conducted on FUNSD to illustrate the impact of various combinations of the pre-training tasks, and the results are concluded in Tab.~\ref{ablation}. By comparing (a) with each of (b) to (e), we find that every pre-training task benefits the proposed ESP. Specifically, (c) and (e) indicate that EE and EL pre-training tasks benefit corresponding downstream tasks most, resulting in an increase of 2.52\% and 27.69\%, respectively. (d) shows that the EITM module indeed learns effective vision-language representation and improves EE and EL by 2.45\% and 0.45\%. By comparing (f) with (g) and (a), we find that the benefits of pre-training tasks far exceed pre-trained data (6.02\% vs 0.06\%). We also experimentally prove that the last three semantic pre-training tasks can boost entity detection. By comparing (f) and (a), we find that the addition of EITM, EE, and EL tasks can improve the accuracy of detection by 4.35\%. It demonstrates that the detection and semantic tasks can be mutually reinforced.

\noindent\textbf{Evaluations under Different Restrictions} We also evaluate the results under different restrictions, which are compliant with various input combinations in testing: (a) ground truth of text boxes and contents, (b) predicted text boxes and ground truth of text contents, (c) predicted text boxes and contents. Here, we re-implement LayoutXLM~\cite{xu2021layoutxlm} and TRIE~\cite{zhang2020trie} and compare them with our ESP on SIBR. LayoutXLM is more suitable for SIBR as it is the multi-lingual variant of LayoutLMv2. TRIE and ESP are both trained in an end-to-end manner. As for LayoutXLM, it is trained with OCR-matched truth and evaluated with ground truth. In order to conduct experiments on the third condition (c), we use Duguang OCR engine\footnote{https://duguang.aliyun.com/} to get character-level locations and contents which are not provided by PaddleOCR. We closely follow the experimental setup of TRIE\footnote{https://github.com/hikopensource/DAVAR-Lab-OCR} and LayoutXLM\footnote{https://github.com/microsoft/unilm}. The ablation studies under different restrictions are shown in~\cref{res_on_SIBR}.

(a) \textit{With GT Input}. ESP achieves SOTA results on both EE and EL tasks, outperforming the other two methods by 0.55\% and 1.97\%, respectively. This proves that ESP can effectively handle the challenges in SIBR.

(b) \textit{With Predicted Boxes}. In the second comparison, the performance of ESP exceeds the other two methods by a large margin. Specifically, taking the detection errors of PaddleOCR into consideration, the performance of LayoutXLM decreases by 19.65\% in EE and 30.36\% in EL. The performance of TRIE decreases by a larger margin. However, ESP only decreases by 2.98\% and 11.52\%, respectively, demonstrating its robustness to detection error. 

(c) \textit{With Predicted Boxes and Text}. Taking the recognition results into consideration, two of the three methods' F1-scores decrease by a significant margin (\textgreater 20\%). Since TRIE gives an egregious result, we leave the corresponding result blank. We have several conclusions in this experiment: (1) The advantages of ESP become smaller with predicted text contents, (2) The text recognition branch is not easily trained with small data, 
(3) The connection between OCR and VIE should be studied more. The ultimate goal of VIE is to get not only the correct entity labels but also the correct contents. A qualitative comparison is shown in Fig.~\ref{fig:final_vis}. Compared with LayoutXLM, ESP can better handle challenges such as overlap and printing shift.

\noindent\textbf{Effect of Text Recognition} We conduct the following experiments to discuss the effectiveness of the optional text recognition branch, and the results are presented in ~\cref{res_on_SIBR}. $\text{ESP}^{\dagger}$ refers to a co-training process of ESP with a text recognition branch. It can be observed from the first comparison that the result of entities slightly decreases by 0.34\% with a recognition branch. The reason for the drop is the insufficient fine-tuning of text recognition on small data. Besides, we presume that the pre-training tasks have learned a better vision-language representation, which means explicit text information is not essential to the proposed ESP.

\section{Conclusion and Future Work}

We have released an in-the-wild document dataset for VIE, in which the images are challenging in their diverse appearances and complex structures. To deal with these difficulties, we further propose a unified framework to model entities as semantic points, in which entity spotting, labeling, and linking can be learned in an end-to-end manner. We also introduce three pre-training tasks to enforce the network to learn a better vision-language representation. Extensive experiments demonstrate that the proposed method can achieve significantly enhanced performance. The proposed ESP can also be extended to more general VIE problems, which we will leave for future research. 

\noindent\textbf{Acknowledgement.}
This work was supported by the National Natural Science Foundation of China 62225603.

{\small
\bibliographystyle{ieee_fullname}
\bibliography{ESP}
}

\end{document}